\pdfoutput=1
\documentclass[11pt]{article}

\usepackage[final]{acl}

\usepackage{times}
\usepackage{latexsym}

\usepackage[T1]{fontenc}
\usepackage[utf8]{inputenc}

\usepackage{microtype}

\usepackage{inconsolata}

\usepackage{graphicx}

\usepackage{amsmath, amssymb}
\usepackage{algpseudocode}
\usepackage{algorithm}
\usepackage{tabularx}
\usepackage{booktabs}
\usepackage{color}
\usepackage{soul}
\usepackage{subcaption}
\usepackage{array}

\usepackage {tablefootnote}
\usepackage[hyphens]{xurl}
\usepackage{comment}

\usepackage{xspace}
\usepackage[normalem]{ulem}

\usepackage{xltabular}

\setul{0.5ex}{0.2ex}
\newcommand{\ulwith}[2]{\raisebox{-1.0ex}{\scriptsize{\textbf{#1}}}\,\ul{\textbf{#2}}}

\newcommand{\ours}{OPTS\xspace}
\newcommand{\TSBase}{OPTS(TS)\xspace}
\newcommand{\RSBase}{OPTS(US)\xspace}
\newcommand{\LLMBase}{OPTS(APET)\xspace}

\newcommand{\set}[1]{\mathcal{#1}}

\newcolumntype{C}[1]{>{\centering\arraybackslash}p{#1}}
\newcolumntype{L}[1]{>{\raggedleft\arraybackslash}p{#1}}

\title{Bandit-Based Prompt Design Strategy Selection\\Improves Prompt Optimizers}

\author{
  \textbf{Rin Ashizawa}
  \enspace
  \textbf{Yoichi Hirose}
  \enspace
  \textbf{Nozomu Yoshinari}
  \\
  \textbf{Kento Uchida}
  \quad
  \textbf{Shinichi Shirakawa}
  \\
  Yokohama National University
  \\
  \texttt{\{ashizawa-rin-gs,hirose-youichi-kc,yoshinari-nozomu-ry\}@ynu.jp} \\
  \texttt{\{uchida-kento-fz,shirakawa-shinichi-bg\}@ynu.ac.jp}
}

\begin{document}
\maketitle
\begin{abstract}
Prompt optimization aims to search for effective prompts that enhance the performance of large language models (LLMs).
Although existing prompt optimization methods have discovered effective prompts, they often differ from sophisticated prompts carefully designed by human experts. 
Prompt design strategies, representing best practices for improving prompt performance, can be key to improving prompt optimization.
Recently, a method termed the Autonomous Prompt Engineering Toolbox (APET) has incorporated various prompt design strategies into the prompt optimization process.
In APET, the LLM is needed to implicitly select and apply the appropriate strategies because prompt design strategies can have negative effects.
This implicit selection may be suboptimal due to the limited optimization capabilities of LLMs.
This paper introduces Optimizing Prompts with sTrategy Selection (OPTS), which implements explicit selection mechanisms for prompt design. We propose three mechanisms, including a Thompson sampling-based approach, and integrate them into EvoPrompt, a well-known prompt optimizer.
Experiments optimizing prompts for two LLMs, Llama-3-8B-Instruct and GPT-4o mini, were conducted using BIG-Bench Hard. Our results show that the selection of prompt design strategies improves the performance of EvoPrompt, and the Thompson sampling-based mechanism achieves the best overall results. Our experimental code is provided at \url{https://github.com/shiralab/OPTS}.
\end{abstract}

\begin{figure*}[t]
  \centering
  \includegraphics[width=0.98\linewidth]{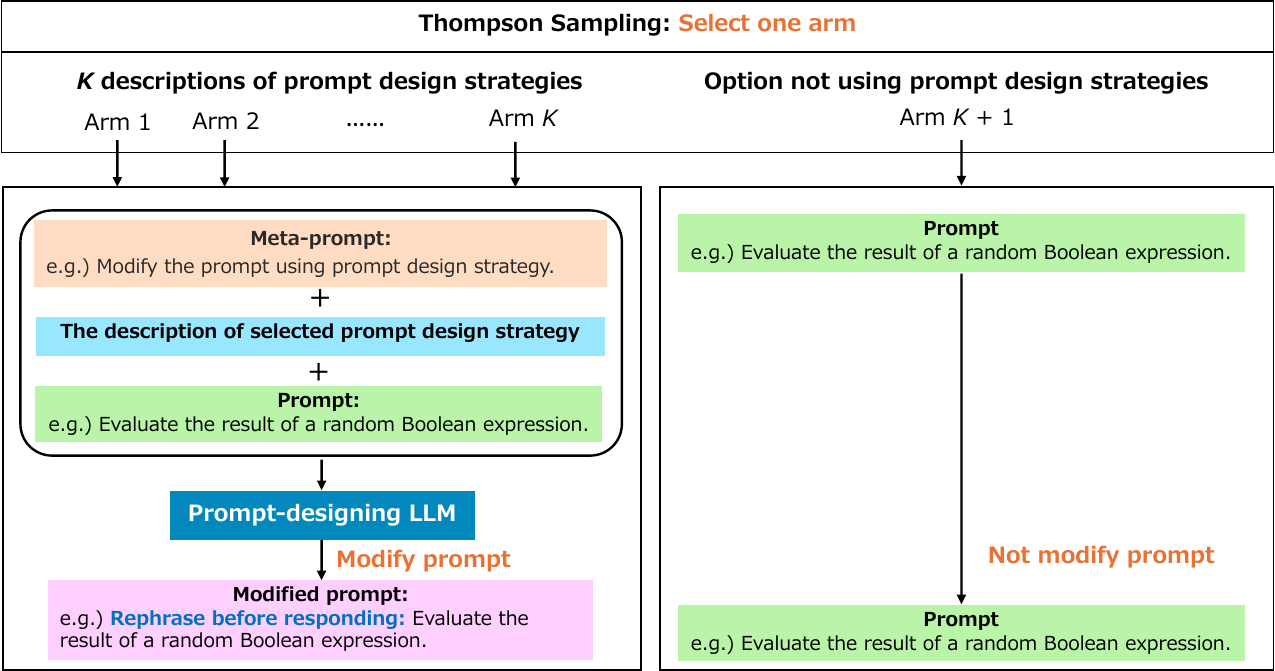}
  \caption{
  Overview of \TSBase, which shows how the prompt generated from the prompt optimizer is modified.
  If one of the first $K$ arms is selected, the description of the prompt design strategy corresponding to the selected arm is passed to the prompt-designing LLM.
  If the $(K+1)$-th arm is selected, the prompt is not modified in any way.
  }
  \label{fig:overview}
\end{figure*}

\section{Introduction}
Large language models (LLMs) such as GPT-4~\citep{openai2024gpt4technicalreport}, Gemini~\citep{geminiteam2024geminifamilyhighlycapable}, and Llama 3~\citep{grattafiori2024llama3herdmodels} have demonstrated superior abilities in a variety of domains, including medicine~\citep{nori2023capabilitiesgpt4medicalchallenge}, law~\citep{doi:10.1098/rsta.2023.0254}, and code generation~\citep{rozière2024codellamaopenfoundation}.
Since well-crafted prompts improve the performance of LLMs, prompt engineering (i.e., designing better prompts) plays a key role in the area~\citep{bsharat2024principledinstructionsneedquestioning, schulhoff2024promptreportsystematicsurvey}.
Despite its importance, prompt engineering is laborious as it requires a lot of time for refinement and sufficient knowledge of the tasks. 
To design effective prompts with less effort, research on prompt optimization has been actively conducted.
In particular, discrete prompt optimization, which optimizes prompts within the natural language space, has attracted attention. This approach is valuable as it typically allows for the optimization of prompts for black-box LLMs, such as GPT-4, while also providing interpretable results~\citep{chang2024efficientpromptingmethodslarge}.
To explore effective prompts within the large natural language space, several methods have been proposed, which include emulating evolutionary algorithms using LLMs~\citep{guo2024connecting, fernando2023promptbreederselfreferentialselfimprovementprompt, cui2024phaseevo}.
These methods have discovered effective prompts, but they often differ from sophisticated prompts carefully designed by human experts.

Prompt design strategies, which provide guidelines for creating effective prompts, can be key to boosting the performance of prompt optimizers.
In fact, Chain-of-Thought (CoT; ~\citealp{NEURIPS2022_9d560961}) and Role Prompting~\citep{wang-etal-2024-rolellm} have been employed in prompt optimization, leading to better prompts~\citep{agarwal2024promptwizardtaskawarepromptoptimization}.
Recently, \citet{kepel2024autonomouspromptengineeringlarge} proposed a method termed the Autonomous Prompt Engineering Toolbox (APET), which incorporated various prompt design strategies into the optimization process.
APET fed all prepared strategies into an LLM to generate a new prompt that incorporates the strategies.
However, not all strategies should be incorporated because prompt design strategies can have negative effects depending on both the LLM and the task~\citep{zheng-etal-2024-helpful, deng2024rephraserespondletlarge}.
In APET, an LLM that generates prompts is required to implicitly select appropriate strategies, which may lead to suboptimal performance because LLMs cannot perform optimization effectively~\citep{huang2024exploringtruepotentialevaluating}.

In this paper, we introduce explicit selection mechanisms for prompt design strategies for the first time.
We also propose three selection methods, including one based on Thompson sampling (TS; \citealp{thompson1933likelihood, russo2018tutorial}).
By integrating them with the existing prompt optimizer, EvoPrompt~\citep{guo2024connecting}, we show that explicit strategy selection effectively leverages existing knowledge of prompt design and enhances the performance of prompt optimizers.
Moreover, the optimizer with the TS-based selection mechanism outperforms other existing methods.

In summary, our contributions are as follows:
\setlength{\leftmargini}{10pt}
\begin{itemize}
    \item We propose explicit prompt design strategy selection mechanisms, including a method based on Thompson sampling, for prompt optimizers.
    \item We experimentally show that the proposed selection mechanism enhances EvoPrompt.
    TS-based selection improves EvoPrompt's performance by up to 50\% when using GPT-4o mini for both generating prompts and solving downstream tasks.
    \item We also compare the TS-based selection with APET-based selection and uniform sampling-based selection.
    The results demonstrate that TS-based selection is overall superior.
\end{itemize}

\section{Related Work}
\paragraph{Prompt design strategy.} The term prompt design strategy refers to a well-established guideline for designing prompts that have been empirically known to be effective.
Chain-of-Thought (CoT; \citealp{NEURIPS2022_9d560961}) and Role Prompting~\citep{wang-etal-2024-rolellm} are notable examples.
CoT asks the LLMs to generate not only the answer, but also the reasoning process that leads to the answer.
Role Prompting is a strategy that includes phrases in the prompt that give the LLM a role.
Various prompt design strategies have been proposed so far~\citep{schulhoff2024promptreportsystematicsurvey,xu2023expertpromptinginstructinglargelanguage,li2023largelanguagemodelsunderstand,xu-etal-2024-reading,deng2024rephraserespondletlarge,lu-etal-2023-bounding,bsharat2024principledinstructionsneedquestioning}, yet they are not always useful.
Indeed, CoT and Role Prompting can lead to worse results~\citep{deng2024rephraserespondletlarge,zheng-etal-2024-helpful}.
As their efficacy depends on the LLM and task, users need to make a non-obvious decision on whether to use them.

\paragraph{Discrete prompt optimization.}
Discrete prompt optimization optimizes prompts in natural language space.
To effectively deal with natural language space, several prompt optimizers emulate the process of black-box optimization algorithms by using LLMs.
These methods are useful because optimized prompts have high interpretability, while they can be applied to LLMs that can be accessed through black-box APIs such as GPT-4~\citep{openai2024gpt4technicalreport}.
G\textsc{r}IPS~\citep{prasad-etal-2023-grips} repeatedly edits the phrases in the prompt, and APE~\citep{zhou2023large} repeatedly generates prompts using LLMs based on Monte Carlo search.
Unlike APE and G\textsc{r}IPS, ProTeGi~\citep{pryzant-etal-2023-automatic} uses a mechanism in which incorrect answers made by an LLM and the corresponding prompt are fed into another LLM to generate a proposal to improve the prompt, and another LLM responsible for designing prompts then modifies the prompt according to the proposal.
In addition to this mechanism, PromptAgent~\citep{wang2024promptagent} also uses Monte Carlo Tree Search to efficiently optimize prompts.
Besides these, adv-ICL~\citep{long-etal-2024-prompt}, which applies adversarial learning, has been proposed.
In OPRO~\citep{yang2024large}, instead of using an LLM to suggest a proposal to improve the prompts, an LLM directly generates new prompts using three items: previously generated prompts, their scores, and a description of the downstream task.

Recently, several methods combining LLMs with evolutionary algorithms have been proposed~\citep{guo2024connecting,jin2024zeroshotchainofthoughtreasoningguided,fernando2023promptbreederselfreferentialselfimprovementprompt,cui2024phaseevo}.
EvoPrompt~\citep{guo2024connecting}, a representative method among them, emulates the optimization process of Genetic Algorithm (GA) or Differential Evolution (DE).
In contrast to EvoPrompt~\citep{guo2024connecting}, PromptBreeder~\citep{fernando2023promptbreederselfreferentialselfimprovementprompt} also optimizes the prompt that is used for generating new prompts.
PhaseEvo~\citep{cui2024phaseevo} optimizes both task instruction and examples and achieves highly effective optimization by dividing optimization into multiple stages.

In addition to dividing optimization into multiple stages, PromptWizard~\citep{agarwal2024promptwizardtaskawarepromptoptimization} utilizes prompt design strategies such as CoT and Role Prompting, but it lacks a strategy selection mechanism and applies them in all cases.
EoT prompting~\citep{jin2024zeroshotchainofthoughtreasoningguided} optimizes zero-shot CoT~\citep{NEURIPS2022_8bb0d291} using evolutionary algorithms.
APET \citep{kepel2024autonomouspromptengineeringlarge} is the most relevant to our study.
In APET, a prompt and descriptions of prompt design strategies are input to an LLM.
The LLM then implicitly selects strategies and generates a new prompt.
In contrast, we propose explicit strategy selection mechanisms that assist prompt optimizers in exploiting appropriate strategies.

\begin{algorithm*}[ht]
\begin{algorithmic}[1]
\Require {Initial prompts $\set{P}_0 = \{p_1, p_2, \dots, p_N\}$, population size $N$, number of iterations $T$, development set $D_{\mathrm{dev}}$ consisting of input and correct output pairs $(x, y)$, scoring function $g$, task-solving LLM $f_T$}
\State \textbf{Evaluation} of initial prompts: $\set{S}_{0} \leftarrow{\left\{s_i=\frac{1}{|D_{\mathrm{dev}}|}\sum_{(x,y) \in D_{\mathrm{dev}}}g\left(y, f_T\left(p_i, x\right)\right) : p_i \in \set{P}_0\right\}}$
\For{$t = 1$ to $T$}
    \For{$p_i$ in $\set{P}_{t-1}$}
        \Comment{$p_i$: the $i$-th parent prompt}
        \State \textbf{Sample donors}: $p_{r_1}, p_{r_2} \in \set{P}_{t-1}$
       , where $p_{r_1}$, $p_{r_2}$, and $p_i$ differ from each other.
        \State \textbf{Crossover and Mutation}: $p_i' \leftarrow f_{D}(m_{\mathrm{de}}, (p_i, p_{r_1}, p_{r_2}, p_{\mathrm{best}}))$ \\ \qquad \quad where $p_{\mathrm{best}}$ is the current best prompt.
        \Comment{$f_{D}$: Prompt-Designing LLM}\\
        \Comment{$m_{\mathrm{de}}$: Meta-prompt for DE-based crossover and mutation}
        \State \textbf{\ours}: Generate $p_i''$ from $p_i'$ by incorporating prompt design strategies (Refer to Section~\ref{sec:OPTS})
        \State \textbf{Selection}:
       $p^{*}_i = \underset{p \in \{p_{i}, p_{i}''\}}{\mathrm{argmax}}\ \frac{1}{|D_{\mathrm{dev}}|}\sum_{(x,y) \in D_{\mathrm{dev}}}g\left(y, f_T\left(p, x\right)\right)$\\
        \Comment{Keep the better one in the population}
        \State \textbf{Update probability distribution} if the TS-based selection is used (Refer to Section~\ref{sec:OPTS})
   \EndFor  
   \State \textbf{Update}: $\set{P}_t \leftarrow \{p_i^* : 1 \le i \le N\}$ 
\EndFor
\State \textbf{Return} the best prompt $p^* = \mathrm{argmax}_{p \in \set{P}_T} \frac{1}{|D_{\mathrm{dev}}|}\sum_{(x,y) \in D_{\mathrm{dev}}}g\left(y, f_T\left(p, x\right)\right)$
\caption{EvoPrompt(DE)-\ours}
\label{alg:DE}
\end{algorithmic}
\end{algorithm*}

\section{Proposed Methods}
In this section, we describe our proposed methods for selecting prompt design strategies.
We then introduce prompt optimization algorithms that combine EvoPrompt~\citep{guo2024connecting} with the strategy selection methods.
We term our methods Optimizing Prompts with sTrategy Selection (OPTS).

\paragraph{Terminology}
\textit{Task-solving LLM} is an LLM that is applied to and solves downstream tasks, while \textit{Prompt-designing LLM} is another LLM that produces helpful prompts for task-solving LLMs.
In contrast to \textit{prompt}, which represents an instruction for a task-solving LLM, \textit{meta-prompt} refers to an instruction for a prompt-designing LLM.

\subsection{Selection of Prompt Design Strategies}
\label{sec:OPTS}
In the following, we discuss three different methods: the TS-based selection called \TSBase, the uniform sampling-based selection called \RSBase, and the APET-based selection called \LLMBase.

\paragraph{\TSBase} \TSBase selects prompt design strategies using Thompson sampling (TS; \citealp{thompson1933likelihood, russo2018tutorial}), which is a multi-armed bandit algorithm and empirically shows superior performance~\citep{NIPS2011_e53a0a29}.

The overview of \TSBase is shown in Figure~\ref{fig:overview}.
There are $K$ arms, each corresponding to one of the $K$ descriptions of the prompt design strategies.
Also, we append a special arm called the \textit{inaction arm}, which corresponds to the option of not using the prompt design strategy, making a total of $K + 1$ arms.
The inaction arm is needed because none of the predefined strategies may improve the prompts at all.
To instantiate TS, we use the beta distributions as priors for the expected reward.
Once one of the first $K$ arms is sampled by TS, we feed the description of the corresponding prompt design strategy into the prompt-designing LLM along with the meta-prompt and the prompt to be modified.
The LLM then modifies the prompt based on the input.
The meta-prompt is the same as that used in APET, whose details are explained in Appendix~\ref{appendix:apet}.
After evaluating the generated prompt with the task-solving LLM and calculating its score, a reward is calculated using the score, and the distributions in TS are updated based on the reward.
Throughout this paper, we compute the reward $r$ for a prompt with the score $s$ as
\begin{equation}
    r = \boldsymbol{1}\left[s > \max \tilde{\set{S}}\right] \in \{0, 1\} \enspace,
\end{equation}
where $\tilde{\set{S}}$ is the set of scores of the parent prompts, which come from EvoPrompt~\cite{guo2024connecting} described in Section~\ref{sec:combined}, and $\boldsymbol{1}[\,\cdot\,]$ is the indicator function.

\paragraph{\RSBase} In \RSBase, each arm is selected according to a uniform distribution.
\RSBase is similar to \TSBase, except that the probability of selecting each arm is equal and is not updated.

\paragraph{\LLMBase} \LLMBase is the selection method based on APET~\citep{kepel2024autonomouspromptengineeringlarge}.
It is slightly different from APET in that it has an additional option equivalent to the inaction arm.
\LLMBase first randomly decides with a probability of 0.5 whether to modify a prompt based on the prompt design strategies.
If it decides to modify the prompt, the prompt-designing LLM is applied to the prompt to incorporate the prompt design strategies.
The prompt-designing LLM receives the meta-prompt, the description of all prompt design strategies, and the prompt to be modified.
Then, it implicitly selects the prompt design strategies and modifies the prompt according to them.

\begin{table*}[t]
    \centering
    \small
    \renewcommand{\arraystretch}{1.2}
    \begin{tabularx}{\linewidth}{p{45mm}X}
        \toprule
        Prompt Design Strategy & Remarks \\
        \midrule
        ExpertPrompting & Assign expert roles to task-solving LLMs~\citep{xu2023expertpromptinginstructinglargelanguage}. \\
        Chain-of-Thought & Let task-solving LLMs also generate a reasoning process~\citep{NEURIPS2022_9d560961}. \\
        Tree-of-Thought &  Let task-solving LLMs iteratively choose the best of multiple reasoning paths, backtracking as necessary~\citep{NEURIPS2023_271db992}. \\
        Emotion Prompting & Incorporate phrases that appeal to human emotions~\citep{li2023largelanguagemodelsunderstand}. \\
        Re-Reading & Instruct task-solving LLMs to reread the question~\citep{xu-etal-2024-reading}. \\
        Style Prompting & Specifies the desired output style~\citep{lu-etal-2023-bounding}. \\
        Rephrase and Respond & Let task-solving LLMs rephrase the question before responding~\citep{deng2024rephraserespondletlarge}. \\
        Avoiding bias & A more generalized version of the 13th principle of the 26 principles of prompting~\citep{bsharat2024principledinstructionsneedquestioning}. \\
        Making prompt specific & Based on Best practices for prompt engineering published by OpenAI.\footnotemark \\
        Shortening the prompt & Based on the experimental result that accuracy can decrease as prompts become longer~\citep{levy-etal-2024-task}.\\
        Adding necessary information & One of the strategies used in APET~\citep{kepel2024autonomouspromptengineeringlarge}.\\
        \bottomrule
    \end{tabularx}
    \caption{Prompt design strategies used in the experiment.
    The concrete descriptions of each strategy are provided in Appendix \ref{appendix:descriptions}.
    }
    \label{table:strategy}
\end{table*}

\subsection{EvoPrompt with OPTS}
\label{sec:combined}
We combine the proposed selection methods with EvoPrompt~\citep{guo2024connecting}.
We adopt EvoPrompt because it is effective yet sufficiently simple, allowing us to focus solely on evaluating the strategy selection methods.
Also, it has variants depending on whether GA or DE is employed.
This feature allows us to assess the impact of prompt design strategy selection on different optimization algorithms.
The algorithm integrated \ours into EvoPrompt(DE) is shown in Algorithm~\ref{alg:DE}, while that based on EvoPrompt(GA) is shown in Appendix~\ref{appendix:algorithm-details}.
Note that a response generation by LLM $f$ is denoted by $f\left(p, x\right)$.
We insert \ours after the crossover and mutation process of EvoPrompt.
After evaluating the newly generated prompts with the task-solving LLM, the scores are used to determine the next generation's population and, if necessary, to update the distribution of the arms in TS.
See Appendix~\ref{appendix:algorithm-details} for the details of the algorithm.

\begin{table*}[t]
  \newcommand{\mysize}{\normalsize}
  \newcommand{\mysizeII}{\small}
  \newcommand{\avgsize}{\large}
  \centering
  \resizebox{\textwidth}{!}{
  \begin{tabular}{C{5mm}p{50mm}C{10mm}C{10mm}C{23mm}C{25mm}C{25mm}C{25mm}}
    \toprule
    Task ID & Task Name & Manual Prompt & APET & EvoPrompt(DE) & EvoPrompt(DE)-\LLMBase & EvoPrompt(DE)-\RSBase & EvoPrompt(DE)-\TSBase \\
    \midrule
    {\mysize 0} & {\mysize boolean expressions} & 54.00 & 67.50 & 74.50 (1.08) & 79.83 (2.05) & \textbf{84.00} (0.41) & 82.50 (4.95) \\
    {\mysize 1} & {\mysize causal judgement} & 2.19 & 0.00 & 40.39 (3.00) & 42.34 (2.98) & 40.15 (4.88) & \textbf{45.50} (3.00) \\
    {\mysize 2} & {\mysize date understanding} & 14.00 & 3.00 & 17.17 (1.03) & 19.17 (4.29) & 18.67 (5.57) & \textbf{20.17} (6.28) \\
    {\mysize 3} & {\mysize disambiguation qa} & 19.50 & 22.50 & 30.00 (1.87) & 38.33 (4.11) & \textbf{47.33} (6.54) & 42.50 (5.31) \\
    {\mysize 4} & {\mysize dyck languages} & 6.50 & 0.00 & 6.67 (0.85) & 6.50 (0.71) & 6.17 (0.85) & \textbf{7.50} (0.00) \\
    {\mysize 5} & {\mysize formal fallacies} & 29.50 & 0.00 & 40.67 (1.31) & \textbf{44.83} (2.66) & 42.50 (1.22) & 43.50 (3.56) \\
    {\mysize 6} & {\mysize geometric shapes} & 16.50 & 24.50 & \textbf{36.00} (0.41) & 33.00 (0.41) & 33.00 (4.32) & 35.83 (2.62) \\
    {\mysize 7} & {\mysize hyperbaton} & 53.00 & 3.50 & 54.67 (0.85) & \textbf{70.00} (0.82) & 59.50 (5.02) & 60.50 (4.42) \\
    {\mysize 8} & {\mysize logical deduction five objects} & 12.00 & 3.50 & 14.67 (1.03) & 29.00 (6.48) & 24.67 (7.15) & \textbf{37.17} (13.21) \\
    {\mysize 9} & {\mysize logical deduction seven objects} & 5.50 & 3.00 & 5.83 (0.24) & 10.17 (1.25) & \textbf{13.17} (0.85) & 13.00 (1.87) \\
    {\mysize 10} & {\mysize logical deduction three objects} & 44.00 & 20.50 & 45.83 (3.42) & 70.17 (5.10) & 71.83 (2.09) & \textbf{78.83} (5.98) \\
    {\mysize 11} & {\mysize movie recommendation} & 40.50 & 0.00 & 49.50 (5.12) & 49.50 (2.83) & 48.00 (2.48) & \textbf{54.33} (1.89) \\
    {\mysize 12} & {\mysize multistep arithmetic two} & \textbf{53.50} & 52.00 & \textbf{53.50} (1.78) & 52.17 (2.46) & 50.33 (0.85) & 52.50 (1.47) \\
    {\mysize 13} & {\mysize navigate} & 84.50 & 38.50 & 83.17 (0.47) & 80.17 (2.66) & 81.67 (3.40) & \textbf{84.67} (1.03) \\
    {\mysize 14} & {\mysize object counting} & \textbf{87.50} & 27.50 & 85.83 (0.62) & 85.67 (0.62) & 85.00 (0.82) & 85.83 (0.24) \\
    {\mysize 15} & {\mysize penguins in a table} & 26.04 & 8.33 & 22.57 (3.93) & 28.47 (7.23) & \textbf{41.67} (5.17) & 40.97 (7.90) \\
    {\mysize 16} & {\mysize reasoning about colored objects} & 21.00 & 6.50 & \textbf{53.00} (3.54) & 50.00 (4.64) & 45.67 (5.27) & 49.50 (3.27) \\
    {\mysize 17} & {\mysize ruin names} & 18.50 & 65.50 & 27.17 (2.90) & \textbf{68.67} (7.26) & 64.83 (2.39) & 67.67 (3.66) \\
    {\mysize 18} & {\mysize salient translation error detection} & 12.50 & 6.50 & 46.17 (3.52) & 46.50 (4.71) & \textbf{51.17} (1.55) & \textbf{51.17} (3.66) \\
    {\mysize 19} & {\mysize snarks} & 24.22 & 0.00 & 55.47 (6.72) & 58.59 (3.31) & \textbf{65.10} (4.25) & 64.06 (1.10) \\
    {\mysize 20} & {\mysize sports understanding} & 51.52 & 0.00 & 61.45 (1.67) & 61.62 (1.43) & 76.94 (7.76) & \textbf{78.45} (4.54) \\
    {\mysize 21} & {\mysize temporal sequences} & 57.00 & 6.00 & 65.83 (3.06) & 65.83 (1.18) & 60.33 (0.62) & \textbf{66.00} (1.47) \\
    {\mysize 22} & {\mysizeII tracking shuffled objects five objects} & 67.50 & 18.00 & 66.50 (2.16) & 66.17 (0.94) & 68.17 (1.25) & \textbf{69.33} (1.25) \\
    {\mysize 23} & {\mysizeII tracking shuffled objects seven objects} & 47.50 & 19.50 & 52.00 (1.22) & 52.50 (1.87) & \textbf{54.33} (1.65) & 49.50 (0.71) \\
    {\mysize 24} & {\mysizeII tracking shuffled objects three objects} & 80.50 & 80.50 & 78.67 (1.55) & 77.50 (2.83) & 79.17 (1.65) & \textbf{81.67} (2.01) \\
    {\mysize 25} & {\mysize web of lies} & 93.50 & 42.00 & \textbf{95.00} (0.00) & \textbf{95.00} (0.00) & \textbf{95.00} (0.00) & \textbf{95.00} (0.00) \\
    {\mysize 26} & {\mysize word sorting} & 44.50 & 41.00 & 45.50 (0.71) & 45.83 (4.11) & \textbf{46.17} (1.70) & 45.33 (2.25) \\
    \midrule
    & {\avgsize AVG} & {\avgsize 39.52} & {\avgsize 20.73} & {\avgsize 48.43} & {\avgsize 52.87} & {\avgsize 53.87} & {\avgsize \textbf{55.67}} \\
    \bottomrule
    \end{tabular}
  }
  \caption{\label{table:main-DE}
    Accuracy on the test %data
    set for 27 tasks from BBH, evaluated with Llama-3-8B-Instruct as the task-solving LLM.
    The scores are averaged over three trials with different seeds.
    The values in parentheses represent the standard deviation.
    The bold scores indicate that the prompt optimized by the method achieved the highest average score.
  }
\end{table*}

\begin{table}[t]
  \centering
  \resizebox{\linewidth}{!}{
    \begin{tabular}{C{15mm}C{25mm}C{30mm}}
    \toprule
    Task ID & EvoPrompt(DE) & EvoPrompt(DE)-\TSBase \\
    \midrule
    8 & 2.67 (1.43) & \textbf{52.33} (11.45) \\
    19 & 79.17 (0.37) & \textbf{79.43} (1.33) \\
    23 & 80.67 (4.37) & \textbf{81.83} (3.47) \\
    \bottomrule
    \end{tabular}
    }
  \caption{\label{table:gpt-4o-mini-DE}
   Accuracy on the test set for three tasks from BBH, evaluated with GPT-4o mini.
   The task IDs are the same as in Table~\ref{table:main-DE}.
   The scores in this table are the average scores over three trials.
    The values in parentheses represent the standard deviation.
  }
\end{table}

\begin{table*}[t]
{
    \newcommand{\mysize}{\normalsize}
    \centering
    \begin{tabularx}{\linewidth}{p{25mm}Xp{10mm}}
    \toprule
    Method & Description & Score \\
    \midrule
    Manual Prompt & {\mysize Select the humorous edit that 'ruins' the input movie or musical artist name.} & 18.50 \\
    \midrule
    EvoPrompt(DE) & {\mysize Decide on the eccentric twist that 'spoils' the name of the movie or music artist.} & 31.00 \\
    \midrule
    APET & {\mysize \ulwith{(1)}{Imagine you are a creative expert in humor and wordplay, skilled at crafting amusing edits that playfully distort movie or musical artist names. Your task is to select the humorous edit that 'ruins' the given input name in a funny and clever way.}

    \ulwith{(2)}{Let's think step-by-step: First}, \ulwith{(3)}{identify the original name provided}. \ulwith{(2)}{Next}, \ulwith{(3)}{brainstorm potential humorous edits that could transform the name into something amusing while maintaining a connection to the original}. \ulwith{(2)}{Finally}, choose the edit that best exemplifies the concept of 'ruining' the name in a \ulwith{(4)}{lighthearted manner}.
    
    \ulwith{(5)}{Read the question again to ensure clarity before proceeding}. \ulwith{(6)}{Remember, your goal is to evoke laughter and joy through your selection!} } & 65.50 \\
    \midrule
    EvoPrompt(DE)-\TSBase & {\mysize \ulwith{(2)}{Let's think step-by-step! First,} carefully \ulwith{(5)}{read the question again} and \ulwith{(3)}{identify a movie title or musician whose name lends itself to a humorous spoof.} \ulwith{(2)}{Next,} \ulwith{(3)}{creatively reimagine that title or name with an absurdly funny twist that maintains the essence of the original while injecting a comedic element}. \ulwith{(2)}{Finally,} present your funniest version clearly, ensuring it is both memorable and entertaining. \ulwith{(6)}{Let your creativity shine through in this process!} } & 70.50 \\
    \bottomrule
    \end{tabularx}
  \caption{\label{table:prompt_analysis}
   Examples of the discovered descriptions (optimized parts within the prompts).
   These achieved the highest score in each method with Llama-3-8B-Instruct as the task-solving LLM and "ruin names" as the task.
   Underlined texts represent (1) ExpertPrompting, (2) Chain-of-Thought, (3) Making prompt specific, (4) Style Prompting, (5) Re-Reading, and (6) Emotion Prompting.
  }
}
\end{table*}

\section{Experiments}
We experimentally evaluate three strategy selection methods we introduce: \TSBase, \RSBase, and \LLMBase.
Combined with EvoPrompt, these methods are applied to various tasks and compared with the existing baseline methods.

\subsection{Dataset}
We evaluate the proposed method using BIG-Bench Hard (BBH; \citealp{suzgun2022challengingbigbenchtaskschainofthought}).
BBH is a collection of the tasks that are challenging for LLMs.
Details of each task can be found in the original BBH paper.
For each task, we randomly sample 50 examples from the test set and use them as the development set, as done in \citep{guo2024connecting}.
The development set is used for evaluating prompts in the optimization process.
At the end of the optimization, the prompt with the highest score on the development set is tested on the test set excluding the development set.

\subsection{Metrics}
We use accuracy as the scoring function.
When evaluating a prompt using task-solving LLMs, answer parts are first extracted from the responses generated by the LLMs. 
The regular expression used in lm-evaluation-harness~\citep{eval-harness} is used to extract the answer parts.
In the regular expression, the parts following ``the answer is '' are extracted.
Then, the scoring function gives a value of 1 if they exactly match the desired responses and 0 otherwise.

\footnotetext{\url{https://help.openai.com/en/articles/6654000-best-practices-for-prompt-engineering-with-the-openai-api}. Accessed: Jan. 31, 2025.}

\subsection{Implementation Details} \label{sec:implementation}
We evaluate the strategy selection methods with DE-based EvoPrompt.
We also evaluate GA-based algorithm, which is presented in Appendix~\ref{appendix:result}.
We use the 3-shot prompts from the original BBH paper~\citep{suzgun2022challengingbigbenchtaskschainofthought}, but we optimize only the task description and leave the examples unchanged.
We conduct two experiments in which Llama-3-8B-Instruct ~\citep{grattafiori2024llama3herdmodels} and GPT-4o mini are used as the task-solving LLMs.
In the experiments using Llama-3-8B-Instruct, we use all 27 tasks of the BBH.
In the experiments using GPT-4o mini, due to the high API cost, we sample only 3 out of the 27 tasks.
We run three trials with different random seeds in both our experiments with Llama-3-8B-Instruct and GPT-4o mini.
We use GPT-4o mini for the prompt-designing LLMs, set the population size to 10, and perform optimization for 50 iterations.
The prompt design strategies we use are listed in Table~\ref{table:strategy}.

\subsection{Initial Task Descriptions}
The initial task descriptions given at the beginning of the optimization are prepared by the following procedure.
First, we select five task descriptions from the 20 prepared descriptions based on their evaluation scores on the development set.
The 20 task descriptions consist of those used in the original BBH paper~\citep{suzgun2022challengingbigbenchtaskschainofthought} and their 19 paraphrases generated by GPT-4o mini.
For paraphrasing, we use the instruction ``Generate 19 variations of the following instruction while keeping the semantic meaning,'' which is a slightly modified version of the meta-prompt created by \citet{zhou2023large}.
Then, we use the 10 task descriptions, consisting of the five selected task descriptions and their respective paraphrases by GPT-4o mini, as the initial task descriptions.
When paraphrasing the selected descriptions, in the same way as \citet{guo2024connecting}, we use the meta-prompt for resampling with the instruction ``Generate a variation of the following instruction while keeping the semantic meaning,'' which is created by \citet{zhou2023large}.

\subsection{Baseline Methods}
We use the manual prompts that use those introduced in the BBH paper, APET, and EvoPrompt as the baseline methods. In APET, we use the APET procedure to incorporate prompt design strategies into task description in manual prompts.

\subsection{Main Result}
Table~\ref{table:main-DE} shows the test scores using Llama-3-8B-Instruct as the task-solving LLM.

\paragraph{Effects of OPTS.} First, we focus on the effect of the explicit selection mechanism in OPTS.
Table~\ref{table:main-DE} shows that all three variants of OPTS increase the average scores of EvoPrompt(DE) by approximately 4.5\% to 7.5\% compared with the naive EvoPrompt(DE).
In particular, for the task ``ruin names'' (task ID 17), all three variants of EvoPrompt(DE)-OPTS outperform EvoPrompt(DE) by about 40\%.
The results indicate that OPTS can improve the prompt optimizer.

\paragraph{Comparing OPTS Variants.} We compare three selection methods: \TSBase, \RSBase, and \LLMBase.
Table~\ref{table:main-DE} shows that, with EvoPrompt(DE), \TSBase outperforms \RSBase by about 1.7\% and \LLMBase by about 2.7\% on average.
This result indicates that \TSBase can select more suitable strategies for task-solving LLMs and tasks.
In addition, \LLMBase is inferior to \RSBase, implying that LLMs have an incorrect bias in selecting prompt design strategies.

\paragraph{Differences between Tasks.} Table~\ref{table:main-DE} also shows that the improvement achieved by strategy selection methods varies depending on the task.
For example, in the tasks of ``ruin names'' (task ID 17) and ``logical deduction three objects'' (task ID 10), EvoPrompt(DE)-\TSBase outperforms EvoPrompt(DE) by approximately 40\% and 30\%, respectively.
On the other hand, OPTS degrades the performance of EvoPrompt(DE) in several tasks.
A possible reason is that effective prompt design strategies differ for each task.
In tasks with significant improvement, the eleven candidate strategies used in the experiment likely include effective options.
In contrast, it may be difficult for tasks with performance degradation to achieve better performance using only the strategies available among the eleven candidates.
Owing to the inaction arm, we note that the performance degradation is insignificant compared with the degree of performance improvement.

\subsection{Results Using Another Task-Solving LLM}
We also evaluated EvoPrompt(DE)-\TSBase using another task-solving LLM, GPT-4o mini, which is one of the most widely used LLMs. We conducted experiments on three randomly chosen tasks from BBH.  
The experimental setup was the same as the experiment using Llama-3-8B-Instruct.
Table~\ref{table:gpt-4o-mini-DE} shows the accuracy of the test set.
We observe that EvoPrompt(DE)-\TSBase outperforms EvoPrompt(DE) in all three tasks.
In particular, for ``logical deduction five objects'' (task ID 8), \TSBase increases the accuracy by approximately 50\%.
This result suggests that OPTS is likely to be effective regardless of the task-solving LLM.

\begin{table*}[t]
    \centering
    \begin{tabularx}{\linewidth}{p{25mm}X}
    \toprule
    Before or After OPTS & Description \\
    \midrule
    Before & Let's think step-by-step. Identify a humorous variation that spoofs the title of a film or music artist, inventing a funny alteration while preserving its original essence. Before providing your answer, ensure full clarity and understanding.\\
    \midrule
    After & Let's think step-by-step. \textbf{First}, identify a film or music artist whose title can be humorously spoofed. \textbf{Next}, brainstorm a funny alteration that captures the essence of the original title while adding a comedic twist. \textbf{Finally}, clearly articulate your humorous variation, ensuring that it maintains the original's core meaning. \\
    \bottomrule
    \end{tabularx}
  \caption{\label{table:prompt_analysis2}
   Example of a prompt where Chain-of-Thought prompting is already used, and Chain-of-Thought prompting is selected again and modified accordingly.
   The task, task-solving LLM, and seed settings are the same as in Table~\ref{table:prompt_analysis}.
  }
\end{table*}

\section{Analysis}
In this section, we analyze \TSBase from two perspectives: the analyses of the discovered task descriptions and the case where a strategy was applied more than once.
Further analyses, including comparison under the same number of API calls, analysis of prompts generated during and after optimization, and analysis of score variance, can be found in the Appendices~\ref{sec:cost_analysis} to \ref{sec:score_var}.

\paragraph{Analysis of Discovered Descriptions.}
Table~\ref{table:prompt_analysis} shows the discovered task descriptions when using Llama-3-8B-Instruct as the task-solving LLMs and ``ruin names'' as the task to be solved.
We can see that the task description discovered by EvoPrompt(DE) does not use any prompt design strategies and has a structure similar to that of a manual prompt, which is used to obtain the initial prompts for EvoPrompt. 
We consider that, although the crossover and mutation performed by the prompt-designing LLM in EvoPrompt can change the phrases in the prompt, it is difficult to change the structure significantly.
In addition, unlike EvoPrompt(DE), EvoPrompt(DE)-\TSBase discovered the task description with various prompt design strategies, including CoT, Re-Reading, Making prompt specific, and Emotion Prompting.
This result means that \TSBase significantly improves the task description using prompt design strategies.
When comparing EvoPrompt(DE)-\TSBase with APET, APET incorporated more strategies while its score was lower than EvoPrompt(DE)-\TSBase.
Implicit selection by APET has the advantage of selecting and incorporating multiple prompt design strategies at once, but it may also include unnecessary strategies.
Furthermore, we observe that the task description of EvoPrompt(DE)-\TSBase uses several prompt design strategies in combination, although \TSBase selects only one prompt design strategy at a time.
Indeed, Re-Reading and Making prompt specific are used within the CoT.
This shows that EvoPrompt(DE)-\TSBase possesses the ability to combine multiple prompt design strategies as well as APET.

\paragraph{Analysis of Repeated Selection of a Strategy.}
During the optimization process of EvoPrompt-\TSBase, we observed that \TSBase sampled a prompt design strategy that had already been incorporated into the prompt, thereby further modifying it.
Table~\ref{table:prompt_analysis2} illustrates a case in which CoT was applied again to a prompt within the same trial as Table~\ref{table:prompt_analysis}.
The example shows that reapplying CoT further aligned the prompt with the strategy.
Also, the feature introduced in the step can be observed in the best prompt in Table~\ref{table:prompt_analysis}.
This example suggests that repeatedly applying the same strategy helps incorporate it more effectively.

\section{Conclusion and Discussion}
We introduced explicit selection mechanisms into prompt optimization to effectively leverage existing knowledge of prompt design.
Experiments have demonstrated that the three methods we introduced improve the performance of the prompt optimizers.
In particular, the method based on Thompson sampling is the best among those we compared.
The prompts discovered by our methods effectively incorporate several prompt design strategies, which EvoPrompt alone was unable to discover.
Our results highlight the importance of leveraging existing knowledge and selecting it explicitly.

\section*{Limitations}
There are four limitations that remain for future research:
(1) We formulated \TSBase as the Bernoulli bandit, but alternative reward formulation may improve optimization performance.
(2) Although \ours can be easily integrated into various prompt optimizers, we have not attempted to introduce it to other optimizers than EvoPrompt.
(3) We used Thompson sampling to select the prompt design strategy but did not evaluate other sophisticated methods, such as contextual bandit algorithms.
(4) The performance of \ours mechanism can vary depending on the prompt design strategy prepared, but this is not clarified in this paper.
These points remain topics for future research.

\section*{Ethical Considerations}
Our method may involve the risk of being used to optimize prompts that generate malicious content, such as malware or fake news, even though this is not the intention of our method.
At present, it is extremely difficult to reduce this risk.
Although our method may be beneficial to some malicious users, we expect that our method can be more beneficial to many other benevolent users.

\section*{Acknowledgments}
This work was partially supported by JSPS KAKENHI Grant Number JP23H00491, JST PREST Grant Number JPMJPR2133, JST Grant Number JPMJPF2203, NEDO JPNP20006, and a grant from the Kanagawa Prefectural Government of Japan.

\bibliography{custom}

\appendix

\section{Meta-Prompt for OPTS}
\label{appendix:apet}
The meta-prompt we used was based on  APET~\citep{kepel2024autonomouspromptengineeringlarge} as shown in Table~\ref{table:meta-prompt:APET}.
We fed this meta-prompt into the prompt-designing LLMs after replacing the \texttt{<strategy>} tag with descriptions of the prompt design strategies and the \texttt{<input>} tag with the prompt to be modified.
For APET and \LLMBase, the \texttt{<strategy>} tag was replaced with the list of $K$ tags from \texttt{<strategy 1>} to \texttt{<strategy K>}, where the description of each prompt design strategy is inserted.

\section{Details of EvoPrompt-\ours}
\label{appendix:algorithm-details}

The algorithm that integrates \ours into EvoPrompt(GA) is shown in Algorithm~\ref{alg:GA}.
In the following, we provide supplementary explanations regarding EvoPrompt processes.

\paragraph{Crossover and Mutation.}The prompt-designing LLM performs
crossover and mutation based on the meta-prompts and prompts fed into them. The meta-prompts for EvoPrompt(GA) and EvoPrompt(DE) are described in Tables~\ref{table:meta-prompt:GA} and \ref{table:meta-prompt:DE}, respectively. 

In EvoPrompt(GA), two prompts are selected as parents using roulette wheel selection. They are fed into the prompt-designing LLM along with the meta-prompt by replacing \texttt{<prompt1>} and \texttt{<prompt2>} in the meta-prompt provided in Table~\ref{table:meta-prompt:GA} with them.
The LLM then generates an offspring prompt according to the meta-prompt. 

In EvoPrompt(DE), four prompts in the current population are used to generate an offspring prompt: two randomly selected prompts $p_{r_1}$ and $p_{r_2}$, the current best prompt $p_{\mathrm{best}}$, and a parent prompt.
Each prompt in the current population is selected once as the parent prompt.
Those four prompts are fed into the prompt-designing LLM by replacing \texttt{<prompt0>}, \texttt{<prompt1>}, \texttt{<prompt2>}, and \texttt{<prompt3>} in Table~\ref{table:meta-prompt:DE} with the parent prompt, $p_{r_1}$, $p_{r_2}$, and $p_{\mathrm{best}}$, respectively.
The LLM then performs crossover and mutation to generate an offspring prompt according to the meta-prompt.

\begin{table*}[ht]
    \resizebox{\linewidth}{!}{
    \begin{tabular}{p{30mm}p{150mm}}
    \toprule
    System prompt & Imagine yourself as an expert in the realm of prompting techniques for LLMs. Your expertise is not just broad, encompassing the entire spectrum of current knowledge on the subject, but also deep, delving into the nuances and intricacies that many overlook. Your job is to reformulate prompts with surgical precision, optimizing them for the most accurate response possible. The reformulated prompt should enable the LLM to always give the correct answer to the question. \\
    \midrule
    User prompt & Your available prompting techniques include, but are not limited to the following:\newline
    \newline
    - <strategy>\newline
    \newline
    Your approach is methodical and analytical, yet creative. You use a mixture of the prompting techniques, making sure you pick the right combination for each instruction. You see beyond the surface of a prompt, identifying the core objectives and the best ways to articulate them to achieve the desired outcomes.\newline
    \newline
    Output instructions:""""\newline
    You should ONLY return the reformulated prompt. Make sure to include ALL information from the given prompt to reformulate.\newline
    """"\newline
    \newline
    Given above information and instructions, reformulate below prompt using the techniques provided: """"\newline
    <input>\newline
    """\\
    \bottomrule
    \end{tabular}
    }
    \caption{
    Meta-prompt for OPTS and APET~\citep{kepel2024autonomouspromptengineeringlarge}.
    When APET or \LLMBase is used, $K$ \texttt{<strategy>} tags (i.e., \texttt{<strategy 1>}, \texttt{<strategy 2>}, \dots, \texttt{<strategy $K$>}) are provided, and each of them is replaced with one prompt design strategy description.
    }
    \label{table:meta-prompt:APET}
\end{table*}

\begin{table*}[th]
    \resizebox{\linewidth}{!}{
    \begin{tabular}{p{30mm}p{150mm}}
    \toprule
    User prompt & Please follow the instruction step-by-step to generate a better prompt.\newline  
    1. Crossover the following prompts to generate a new prompt:\newline
    Prompt 1: Your task is to classify the comment as one of the following categories: terrible, bad, okay, good, great.\newline
    Prompt 2: In this task, you are given sentences from movie reviews. The task is to classify a sentence as one of the following categories: terrible, bad, okay, good, great.\newline
    2. Mutate the prompt generated in Step 1 and generate a final prompt bracketed with <prompt> and </prompt>.\newline
    \newline
    1. Crossover Prompt: In this task, you are given comments from movie reviews. Your task is to classify each comment as one of the following categories: terrible, bad, okay, good, great.\newline
    2. <prompt>Given a sentence from a movie review, classify it into one of the following categories: terrible, bad, okay, good, or great.</prompt>\newline
    \newline
    Please follow the instruction step-by-step to generate a better prompt.\newline
    1. Crossover the following prompts and generate a new prompt:\newline
    Prompt 1: <prompt1>\newline
    Prompt 2: <prompt2>\newline
    2. Mutate the prompt generated in Step 1 and generate a final prompt bracketed with <prompt> and </prompt>.\newline
    \newline
    1. \\
    \bottomrule
    \end{tabular}
    }
    \caption{
    Meta-prompt for crossover and mutation in EvoPrompt(GA)~\citep{guo2024connecting}.  
    }
    \label{table:meta-prompt:GA}
\end{table*}

\begin{table*}[th]
    \resizebox{\linewidth}{!}{
    \begin{tabular}{p{30mm}p{150mm}}
    \toprule
    User prompt & Please follow the instruction step-by-step to generate a better prompt.\newline
    1. Identify the different parts between the Prompt 1 and Prompt 2:\newline
    Prompt 1: Rewrite the input text into simpler text.\newline
    Prompt 2: Rewrite my complex sentence in simpler terms, but keep the meaning.\newline
    2. Randomly mutate the different parts\newline
    3. Combine the different parts with Prompt 3, selectively replace it with the different parts in step 2 and generate a new prompt.\newline
    Prompt 3: Rewrite the given input text into simpler English sentences while preserving the same meaning, so it can be understood by non-native English speakers.\newline
    4. Crossover the prompt in the step3 with the following basic prompt and generate a final prompt bracketed with <prompt> and </prompt>:\newline
    Basic Prompt: Make the sentence easier for people who do not speak English fluently to comprehend.\newline
    \newline
    1. Identifying the different parts between Prompt 1 and Prompt 2:\newline
    Prompt 1: Rewrite the input text into simpler text.\newline
    Prompt 2: Rewrite my complex sentence in simpler terms, but keep the meaning.\newline
    Different parts:\newline
    "input text" vs "my complex sentence"\newline
    "simpler text" vs "simpler terms, but keep the meaning"\newline
    \newline
    2. Randomly mutate the different parts:\newline
    "input text" -> "provided text"\newline
    "my complex sentence" -> "the difficult sentence"\newline
    "simpler text" -> "easier language"\newline
    "simpler terms, but keep the meaning" -> "simpler words while maintaining the meaning"\newline
    \newline
    3. Combine the different parts with Prompt 3, selectively replace it with the different parts in step 2 and generate a new prompt:\newline
    Prompt 3: Rewrite the given input text into simpler English sentences while preserving the same meaning, so it can be understood by non-native English speakers.\newline
    New Prompt: Transform the provided text into easier language while maintaining the meaning, making it accessible for non-native English speakers.\newline
    \newline
    4. Crossover the prompt in step 3 with the following basic prompt and generate a final prompt bracketed with <prompt> and </prompt>:\newline
    Basic Prompt: Make the sentence easier for people who do not speak English fluently to comprehend.\newline
    Final Prompt: <prompt>Convert the difficult sentence into simpler words while preserving the meaning, so it's easier for non-native English speakers to understand.</prompt>\newline
    \newline \newline
    Please follow the instruction step-by-step to generate a better prompt.\newline
    1. Identify the different parts between the Prompt 1 and Prompt 2:\newline
    Prompt 1: <prompt1>\newline
    Prompt 2: <prompt2>\newline
    2. Randomly mutate the different parts\newline
    3. Combine the different parts with Prompt 3, selectively replace it with the different parts in step2 and generate a new prompt.\newline
    Prompt 3: <prompt3>\newline
    4. Crossover the prompt in the step3 with the following basic prompt and generate a final prompt bracketed with <prompt> and </prompt>:\newline
    Basic Prompt: <prompt0>\newline
    \newline
    1. """\\
    \bottomrule
    \end{tabular}
    }
    \caption{
    Meta-prompt for crossover and mutation in EvoPrompt(DE)~\citep{guo2024connecting}.  
    }
    \label{table:meta-prompt:DE}
\end{table*}

\begin{algorithm*}[th]
\begin{algorithmic}[1]
    \Require {Initial prompts $\set{P}_0 = \{p_1, p_2, \dots, p_N\}$, population size $N$, number of iterations $T$, development set $D_{\mathrm{dev}}$ consisting of input and correct output pairs $(x, y)$, scoring function $g$, task-solving LLM $f_T$}
    \State \textbf{Evaluation} of initial prompts: $\set{S}_{0} \leftarrow{\left\{s_i=\frac{1}{|D_{\mathrm{dev}}|}\sum_{(x,y) \in D_{\mathrm{dev}}}g\left(y, f_T\left(p_i, x\right)\right) : p_i \in \set{P}_0\right\}}$
    \For{$t = 1$ to $T$}
    \For{$i=1$ to $N$}
    \State \textbf{Sampling parents} by roulette wheel: $p_{r_1}, p_{r_2} \in \set{P}_{t-1}$
    \State \textbf{Crossover and Mutation}: $p_i' \leftarrow f_{D}(m_{\mathrm{ga}}, (p_{r_1}, p_{r_2}))$ \\ \Comment{$f_{D}$: prompt-designing LLM}
    \\ \Comment{$m_{\mathrm{ga}}$: Meta-prompt for GA-based crossover and mutation}
    \State \textbf{\ours}: Generate $p_i''$ from $p_i'$ by incorporating prompt design strategies (Refer to Section~\ref{sec:OPTS})
    \State \textbf{Evaluation}: $s_i'' \leftarrow \frac{1}{|D_{\mathrm{dev}}|}\sum_{(x,y) \in D_{\mathrm{dev}}}g\left(y, f_T\left(p_i'', x\right)\right)$
    \State \textbf{Update probability distribution} if the TS-based selection is used (Refer to Section~\ref{sec:OPTS})
    \EndFor  
    \State $\set{\hat{S}}_{t}\leftarrow\{s_i'': 1 \le i \le N\}$, $\set{\hat{P}}_t\leftarrow\{p_i'': 1 \le i \le N\}$
    \State \textbf{Update score}: $\set{S}_t \leftarrow \text{select the best $N$ scores in }\set{S}_{t-1} \cup \set{\hat{S}}_{t}$
    \State \textbf{Update}: $\set{P}_t \leftarrow \text{select the best $N$ prompts in }\set{P}_{t-1} \cup \set{\hat{P}}_{t}$ according to $\set{S}_{t-1} \cup \set{\hat{S}}_{t}$,
    \EndFor
    \State \textbf{Return} the best prompt $p^* =  \mathrm{argmax}_{p \in \set{P}_T} \frac{1}{|D_{\mathrm{dev}}|}\sum_{(x,y) \in D_{\mathrm{dev}}}g\left(y, f_T\left(p, x\right)\right)$
\caption{EvoPrompt(GA)-\ours}
\label{alg:GA}
\end{algorithmic}
\end{algorithm*}

\begin{table*}[thbp]
    \resizebox{\linewidth}{!}{
    \begin{tabular}{p{50mm}p{130mm}}
    \toprule
    ExpertPrompting  & Crafting an expert who is an expert at the given task, by writing a high-quality description about the most capable and suitable agent to answer the instruction in second person perspective. \\
    \midrule
    Chain-of-Thought  & Explaining step-by-step how the problem should be tackled, and making sure the model explains step-by-step how it came to the answer. You can do this by adding "Let's think step-by-step". \\
    \midrule
    Tree-of-Thought & Imagining three different experts who are discussing the problem at hand. All experts will write down 1 step of their thinking, then share it with the group. Then all experts will go on to the next step, etc. If any expert realises they're wrong at any point then they leave. \\
    \midrule
    Adding necessary information & Making sure all information needed is in the prompt, adding where necessary but making sure the question remains having the same objective. \\
    \midrule
    Re-Reading & For a given prompt, add a phrase such as "Read the question again" that instructs the Large Language Models to reread the question before generating an answer. This strategy is particularly effective for complex tasks and helps enhance the quality and reliability of the model's outputs. \\
    \midrule
    Style Prompting & Clearly define the desired style in the given prompt. For example, you might say, "Write a formal letter about..." or "Create a casual conversation discussing...". This guidance helps the model produce text that matches the requested stylistic elements, whether it's formal, informal, technical, or poetic. \\
    \midrule
    Rephrase and Respond & For a given prompt, add a phrase that instructs the Large Language Models to rephrase the question before responding, such as "Rephrase and expand the question, and respond. \\
    \midrule
    Making prompt specific & Make the description of the given prompt more specific. This makes it easier for Large Language Models to correctly execute prompt instructions. \\
    \midrule
    Avoiding bias & To allow Large Language Models to make logical and unbiased inferences, add phrases to a given prompt that instruct it to remove opinionated content. This helps the model concentrate on providing responses based on careful analysis and logical reasoning, minimizing biases. \\
    \midrule
    Shortening the prompt & If a given prompt has long instructions, make it shorter by condensing it to only the essential parts. \\
    \midrule
    Emotion Prompting & At the end of the prompt, add a phrase that evokes a strong emotion. When doing so, keep the following four points in mind:\newline
    1. Define emotional goals: Identify the emotional response you want to evoke, such as encouragement, motivation, or reassurance.\newline
    2. Use positive language: Incorporate words and phrases that are positive and supportive. Examples include "believe in your abilities," "excellent," "success," and "outstanding achievements".\newline
    3. Emphasize key words: Use techniques like exclamation marks and capitalized words to highlight important aspects and to enhance the emotional impact.\newline
    4. Incorporate social and self-esteem cues: Design stimuli that leverage social influence (e.g., group membership, others' opinions) and boost self-esteem and motivation. This can help regulate the emotional response of the Large Language Models and tap into intrinsic motivation. \\
    \bottomrule
    \end{tabular}
    }
    \caption{Descriptions of each prompt design strategy. 
    Descriptions of ExpertPrompting, Chain-of-Thought, Tree-of-Thought, and Adding necessary information are quoted from APET~\citep{kepel2024autonomouspromptengineeringlarge}. The descriptions Re-Reading, Style Prompting, Rephrase and Respond, and Emotion Prompting are modified versions of descriptions created using GPT-4o from \citet{xu-etal-2024-reading}, \citet{lu-etal-2023-bounding}, \citet{deng2024rephraserespondletlarge}, and \citet{li2023largelanguagemodelsunderstand}, respectively.
    Description of Making prompt specific is created by us and is inspired by OpenAI’s prompt engineering best practice.
    Description of Avoiding bias is created by us by referring to and generalizing the 13th principle of the 26 prompting principles proposed in \citet{bsharat2024principledinstructionsneedquestioning}.
    Description of Shortening the prompt was created by us.
    }
    \label{table:strategy_description}
\end{table*}

\section{Descriptions of Prompt Design Strategies}
\label{appendix:descriptions}

Table \ref{table:strategy_description} provides the descriptions for each of the 11 prompt design strategies used in our experiment.

\begin{table*}[th]
  \newcommand{\mysize}{\normalsize}
  \newcommand{\mysizeII}{\small}
  \newcommand{\avgsize}{\large}
  \centering
  \resizebox{\textwidth}{!}{
  \begin{tabular}{C{5mm}p{50mm}C{10mm}C{10mm}C{23mm}C{25mm}C{25mm}C{25mm}}
    \toprule
    Task ID & Task name & Manual prompt & APET & EvoPrompt(GA) & EvoPrompt(GA)-\LLMBase & EvoPrompt(GA)-\RSBase & EvoPrompt(GA)-\TSBase \\
    \midrule
    {\mysize 0} & {\mysize boolean expressions} & 54.00 & 67.50 & 72.83 (3.30) & 84.33 (3.66) & 83.00 (2.94) & \textbf{85.67} (1.65) \\
    {\mysize 1} & {\mysize causal judgement} & 2.19 & 0.00 & 37.71 (1.82) & 40.88 (2.60) & \textbf{44.53} (3.10) & \textbf{44.53} (0.60) \\
    {\mysize 2} & {\mysize date understanding} & 14.00 & 3.00 & 13.67 (0.85) & 14.50 (5.10) & \textbf{21.17} (10.38) & 16.00 (0.82) \\
    {\mysize 3} & {\mysize disambiguation qa} & 19.50 & 22.50 & 32.17 (4.33) & 37.00 (2.86) & 42.67 (3.27) & \textbf{49.67} (6.91) \\
    {\mysize 4} & {\mysize dyck languages} & 6.50 & 0.00 & \textbf{6.67} (0.94) & 6.50 (0.71) & 5.83 (0.24) & 6.50 (0.00) \\
    {\mysize 5} & {\mysize formal fallacies} & 29.50 & 0.00 & 42.50 (0.71) & \textbf{45.17} (2.05) & 43.83 (0.85) & 44.50 (0.82) \\
    {\mysize 6} & {\mysize geometric shapes} & 16.50 & 24.50 & 29.83 (7.85) & \textbf{37.33} (6.14) & 33.00 (4.26) & 32.67 (2.95) \\
    {\mysize 7} & {\mysize hyperbaton} & 53.00 & 3.50 & 54.83 (0.94) & 61.67 (1.55) & 57.67 (3.57) & \textbf{63.67} (6.51) \\
    {\mysize 8} & {\mysize logical deduction five objects} & 12.00 & 3.50 & 12.33 (1.31) & 20.00 (1.22) & 24.67 (4.03) & \textbf{25.83} (7.96) \\
    {\mysize 9} & {\mysize logical deduction seven objects} & 5.50 & 3.00 & 6.83 (3.01) & 7.83 (3.09) & \textbf{12.17} (1.55) & 11.33 (1.70) \\
    {\mysize 10} & {\mysize logical deduction three objects} & 44.00 & 20.50 & 58.17 (5.72) & 64.00 (3.89) & 59.67 (6.02) & \textbf{66.67} (3.30) \\
    {\mysize 11} & {\mysize movie recommendation} & 40.50 & 0.00 & 48.67 (1.18) & 49.67 (4.17) & \textbf{53.50} (4.90) & 52.50 (1.63) \\
    {\mysize 12} & {\mysize multistep arithmetic two} & \textbf{53.50} & 52.00 & 50.33 (2.25) & 49.83 (1.25) & 46.83 (2.72) & 50.17 (1.03) \\
    {\mysize 13} & {\mysize navigate} & \textbf{84.50} & 38.50 & 79.67 (4.25) & 80.17 (2.32) & 83.83 (3.01) & 82.33 (2.25) \\
    {\mysize 14} & {\mysize object counting} & \textbf{87.50} & 27.50 & 85.50 (0.71) & 82.67 (3.70) & 86.67 (0.62) & 84.33 (0.62) \\
    {\mysize 15} & {\mysize penguins in a table} & 26.04 & 8.33 & 25.00 (1.47) & 29.86 (8.26) & 30.90 (2.99) & \textbf{32.29} (1.47) \\
    {\mysize 16} & {\mysize reasoning about colored objects} & 21.00 & 6.50 & 50.17 (9.33) & 47.50 (1.08) & 47.00 (6.38) & \textbf{52.33} (2.72) \\
    {\mysize 17} & {\mysize ruin names} & 18.50 & 65.50 & 51.00 (13.83) & 63.17 (3.52) & \textbf{67.33} (0.24) & 66.83 (0.47) \\
    {\mysize 18} & {\mysize salient translation error detection} & 12.50 & 6.50 & 31.67 (9.74) & 49.00 (4.42) & 48.17 (4.48) & \textbf{52.67} (1.03) \\
    {\mysize 19} & {\mysize snarks} & 24.22 & 0.00 & 45.83 (6.03) & 52.60 (15.94) & \textbf{64.84} (2.92) & 60.16 (4.42) \\
    {\mysize 20} & {\mysize sports understanding} & 51.52 & 0.00 & 62.63 (0.00) & 62.63 (0.00) & \textbf{65.15} (2.58) & 61.95 (6.20) \\
    {\mysize 21} & {\mysize temporal sequences} & 57.00 & 6.00 & \textbf{63.67} (3.70) & 59.33 (2.49) & 61.83 (1.43) & 61.33 (2.78) \\
    {\mysize 22} & {\mysizeII tracking shuffled objects five objects} & 67.50 & 18.00 & 67.67 (0.47) & 67.83 (0.24) & \textbf{68.67} (2.66) & 67.33 (1.03) \\
    {\mysize 23} & {\mysizeII tracking shuffled objects seven objects} & 47.50 & 19.50 & \textbf{53.33} (2.49) & 51.50 (0.82) & 52.00 (1.08) & 50.33 (1.84) \\
    {\mysize 24} & {\mysizeII tracking shuffled objects three objects} & 80.50 & 80.50 & 81.00 (0.71) & 81.67 (0.62) & 80.67 (0.24) & \textbf{82.33} (0.62) \\
    {\mysize 25} & {\mysize web of lies} & 93.50 & 42.00 & \textbf{96.33} (1.03) & 95.83 (0.62) & \textbf{96.33} (1.43) & 96.00 (1.47) \\
    {\mysize 26} & {\mysize word sorting} & 44.50 & 41.00 & \textbf{46.83} (1.65) & 43.67 (4.40) & 46.00 (7.08) & 44.17 (1.03) \\
    \midrule
    & {\avgsize AVG} & {\avgsize 39.52} & {\avgsize 20.73} & {\avgsize 48.40} & {\avgsize 51.34} & {\avgsize 52.89} & {\avgsize \textbf{53.48}} \\
    \bottomrule
    \end{tabular}
  }
  \caption{\label{table:main-GA}
    Accuracy on the test %data
    set for 27 tasks from BBH, evaluated with Llama-3-8B-Instruct as the task-solving LLM.
    The scores are averaged over three trials with different seeds.
    The values in parentheses represent the standard deviation.
    The bold scores indicate that the prompt optimized by the method achieved the highest average score.
  }
\end{table*}

\begin{table*}[th]
  \centering
    \begin{tabular}{C{15mm}C{25mm}C{30mm}}
    \toprule
    Task ID & EvoPrompt(GA) & EvoPrompt(GA)-\TSBase \\
    \midrule
    8 & 1.67 (0.85) & \textbf{48.17} (5.14) \\
    19 & \textbf{79.43} (1.33) & 78.65 (1.84) \\
    23 & 75.67 (1.25) & \textbf{88.33} (1.55) \\
    \bottomrule
    \end{tabular}
    %}
  \caption{\label{table:gpt-4o-mini-GA}
   Accuracy on the test set for three tasks from BBH, evaluated with GPT-4o mini.
   The task IDs are the same as in Table~\ref{table:main-DE}.
   The scores in the table are the average scores over three trials.
    The values in parentheses represent the standard deviation. 
  }
\end{table*}

\section{GA-Based Experimental Results}
\label{appendix:result}
In this section, we present the GA-based experimental results.
Table~\ref{table:main-GA} shows the result of the experiment using Llama-3-8B-Instruct as a task-solving LLM, and Table~\ref{table:gpt-4o-mini-GA} shows the result of the experiment using GPT-4o mini as a task-solving LLM.
Tables~\ref{table:main-GA} and \ref{table:gpt-4o-mini-GA} show that, as in the DE-based experiments, the performance of EvoPrompt (GA) is enhanced by selecting suitable prompt design strategy using OPTS.
In addition, when comparing the variations of OPTS, we observe a tendency that \TSBase is overall the best, followed by \RSBase and \LLMBase, as in the case of the EvoPrompt(DE).
In terms of the extent of improvement, combining the mechanism for selecting a prompt design strategy with EvoPrompt(DE) has a greater impact than that with EvoPrompt(GA).
These results suggest that, although the extent of improvement varies slightly depending on the evolutionary algorithm used in the prompt optimizer, the mechanism to select the prompt design strategy can enhance the prompt optimizer regardless of the evolutionary algorithm used.

% Below are the additions to the revised version.
\begin{table*}[t]
  \newcommand{\mysize}{\normalsize}
  \newcommand{\mysizeII}{\small}
  \newcommand{\avgsize}{\large}
  \centering
  \resizebox{\textwidth}{!}{
  \begin{tabular}{C{5mm}p{60mm}C{25mm}C{25mm}C{25mm}C{25mm}}
    \toprule
    Task ID & Task Name & EvoPrompt(GA) & EvoPrompt(GA)-\TSBase & EvoPrompt(DE) & EvoPrompt(DE)-\TSBase \\
    \midrule
    0 & boolean expressions & 72.83 (3.30) & \textbf{80.50} (2.48) & 74.50 (1.08) & \textbf{82.67} (2.39) \\
    1 & causal judgement & 37.71 (1.82) & \textbf{44.28} (0.69) & 40.39 (3.00) & \textbf{42.82} (2.26) \\
    2 & date understanding & 13.67 (0.85) & \textbf{15.50} (0.41) & \textbf{17.17} (1.03) & 15.17 (1.03) \\
    3 & disambiguation qa & 32.17 (4.33) & \textbf{41.67} (6.36) & 30.00 (1.87) & \textbf{40.83} (4.52) \\
    4 & dyck languages & \textbf{6.67} (0.94) & 6.50 (0.00) & 6.67 (0.85) & \textbf{7.33} (0.24) \\
    5 & formal fallacies & 42.50 (0.71) & \textbf{43.67} (0.62) & 40.67 (1.31) & \textbf{43.50} (3.56) \\
    6 & geometric shapes & 29.83 (7.85) & \textbf{32.17} (3.17) & 36.00 (0.41) & \textbf{36.17} (2.49) \\
    7 & hyperbaton & 54.83 (0.94) & \textbf{57.67} (2.90) & 54.67 (0.85) & \textbf{57.33} (0.24) \\
    8 & logical deduction five objects & 12.33 (1.31) & \textbf{23.83} (2.62) & 14.67 (1.03) & \textbf{32.17} (10.60) \\
    9 & logical deduction seven objects & 6.83 (3.01) & \textbf{11.17} (1.55) & 5.83 (0.24) & \textbf{10.50} (3.19) \\
    10 & logical deduction three objects & 58.17 (5.72) & \textbf{67.33} (9.78) & 45.83 (3.42) & \textbf{77.17} (5.95) \\
    11 & movie recommendation & 48.67 (1.18) & \textbf{52.50} (1.63) & 49.50 (5.12) & \textbf{51.50} (2.86) \\
    12 & multistep arithmetic two & \textbf{50.33} (2.25) & 48.50 (3.89) & \textbf{53.50} (1.78) & 52.50 (1.47) \\
    13 & navigate & 79.67 (4.25) & \textbf{81.67} (1.31) & \textbf{83.17} (0.47) & 82.50 (2.83) \\
    14 & object counting & \textbf{85.50} (0.71) & \textbf{85.50} (1.08) & 85.83 (0.62) & \textbf{86.17} (0.24) \\
    15 & penguins in a table & 25.00 (1.47) & \textbf{30.21} (1.47) & 22.57 (3.93) & \textbf{30.90} (6.44) \\
    16 & reasoning about colored objects & 50.17 (9.33) & \textbf{51.33} (3.66) & \textbf{53.00} (3.54) & 48.00 (1.47) \\
    17 & ruin names & 51.00 (13.83) & \textbf{65.17} (2.62) & 27.17 (2.90) & \textbf{68.17} (2.25) \\
    18 & salient translation error detection & 31.67 (9.74) & \textbf{52.17} (0.24) & 46.17 (3.52) & \textbf{51.83} (2.72) \\
    19 & snarks & 45.83 (6.03) & \textbf{60.16} (7.99) & \textbf{55.47} (6.72) & 55.21 (3.51) \\
    20 & sports understanding & 62.63 (0.00) & \textbf{64.81} (3.10) & 61.45 (1.67) & \textbf{80.81} (7.56) \\
    21 & temporal sequences & \textbf{63.67} (3.70) & 60.67 (2.66) & \textbf{65.83} (3.06) & 64.67 (1.84) \\
    22 & tracking shuffled objects five objects & \textbf{67.67} (0.47) & 66.33 (0.47) & 66.50 (2.16) & \textbf{69.33} (1.25) \\
    23 & tracking shuffled objects seven objects & \textbf{53.33} (2.49) & 51.00 (0.41) & \textbf{52.00} (1.22) & 51.17 (2.49) \\
    24 & tracking shuffled objects three objects & 81.00 (0.71) & \textbf{81.50} (0.41) & 78.67 (1.55) & \textbf{81.67} (2.01) \\
    25 & web of lies & \textbf{96.33} (1.03) & 95.00 (0.00) & \textbf{95.00} (0.00) & \textbf{95.00} (0.00) \\
    26 & word sorting & 46.83 (1.65) & \textbf{47.83} (1.93) & 45.50 (0.71) & \textbf{49.83} (0.94) \\
    \midrule
    & AVG & 48.40 & \textbf{52.54} & 48.43 & \textbf{54.26} \\
    \bottomrule
    \end{tabular}
  }
  \caption{\label{table:cost}
    Accuracy on the test set for 27 tasks from BBH, evaluated with Llama-3-8B-Instruct as the task-solving LLM.
    The prompts with the highest score on the development set are evaluated on the test set, after 500 API calls for GPT-4o mini.
    The scores are averaged over three trials with different seeds.
    The values in parentheses represent the standard deviation.
    The bold scores indicate that the prompt optimized by the method achieved the highest average score.
  }
\end{table*}

\section{Cost Analysis}
\label{sec:cost_analysis}
We compare the performances of EvoPrompt-\TSBase and EvoPrompt under the same number of API calls for GPT-4o mini. Table \ref{table:cost} shows the scores on the test set for prompts with the highest performance on the development set under the constraint of 500 API calls. We use GPT-4o mini as the prompt-designing LLM and Llama-3-8B-Instruct as the task-solving LLM. Table \ref{table:cost} explains that EvoPrompt-\TSBase outperforms EvoPrompt under the same number of API calls. This result means that the proposed method is more cost-effective.

\begin{table*}[t]
{
    \newcommand{\mysize}{\normalsize}
    \centering
    \begin{tabularx}{\linewidth}{p{15mm}Xp{20mm}}
    \toprule
    Iteration & Generated description & Validation 
    Score \\
    \midrule
    3 & "Read the question again: Evaluate the answer of an arbitrary Boolean expression carefully, ensuring all logical operations are considered." & 0.58 \\
    \midrule
    6 & Rephrase and expand the question, and respond: "Assume the role of a specialist in Boolean algebra with a strong background in assessing Boolean expressions. Determine the value of a randomly generated Boolean expression by analyzing the components step-by-step. Please explain your reasoning clearly while considering all relevant variables and logical operations to arrive at the final outcome." & 0.12 \\
    \midrule
    50 (final) & Imagine three specialists in Boolean analysis engaged in a structured discussion to analyze the outcome of a randomly generated Boolean expression step-by-step. Each specialist will first note and record the distinct elements of the expression individually. After this initial step, they will share their insights with one another and collaboratively analyze each element in sequence, providing detailed explanations of their rationale at each stage. If any specialist realizes they have made an error at any point, they will withdraw from the discussion. Ultimately, they will integrate their insights to arrive at the final result of the Boolean expression. & 0.88 \\
    \bottomrule
    \end{tabularx}
  \caption{\label{table:prompt_during_optimization}
   Examples of the generated descriptions (optimized parts within the prompts) during prompt optimization.
   These descriptions are generated by EvoPrompt(DE)-\TSBase with Llama-3-8B-Instruct as the task-solving LLM and ``boolean expressions'' as the task.
  }
}
\end{table*}

\section{Analysis of Prompts Generated During and After Optimization}
We analyze the generated task descriptions during and after prompt optimization. We confirm that different strategies are used in different tasks, and not all prompt design strategies are used. We also observe that the order and way in which the strategies are incorporated are different, even if the same combination of strategies is chosen.
For instance, we pick up the prompts obtained during the optimization for the task ``boolean expressions.'' The generated prompts are listed in Table~\ref{table:prompt_during_optimization}. In the third and sixth iterations, the prompts incorporated the strategies called ``Re-Reading'' and ``Rephrase and Respond'' were generated, respectively. However, the scores of these prompts on the validation dataset were worse than that of the final best prompt. In addition, both prompt design strategies have not remained in the final best prompt. In the final iteration, the Tree-of-Thought strategy was adopted, contrasting with the result observed for another task in Table \ref{table:prompt_analysis}. From these observations, we conclude that suitable strategies for the task were selected during the optimization process.

\begin{table*}[t]
  \centering
  \resizebox{\textwidth}{!}{
  \begin{tabular}{p{65mm}C{25mm}C{25mm}C{25mm}}
    \toprule
    Method & Avg Min & Avg Median & Avg Max \\
    \midrule
    EvoPrompt(DE) & 45.95 & 48.72 & 50.62 \\
    EvoPrompt(DE)-OPTS(APET) & 49.09 & 53.59 & 55.93 \\
    EvoPrompt(DE)-OPTS(US) & 50.45 & 53.77 & 57.39 \\
    EvoPrompt(DE)-OPTS(TS) & 51.83 & 55.93 & 59.24 \\
    \bottomrule
    \end{tabular}
  }
  \caption{\label{table:variance-DE}
    Averages of the minimum, median, and maximum scores in three trials with DE-based EvoPrompt. 
  }
\end{table*}

\begin{table*}[t]
  \centering
  \resizebox{\textwidth}{!}{
  \begin{tabular}{p{65mm}C{25mm}C{25mm}C{25mm}}
    \toprule
    Method & Avg Min & Avg Median & Avg Max \\
    \midrule
    EvoPrompt(GA) & 44.37 & 48.86 & 51.98 \\
    EvoPrompt(GA)-OPTS(APET) & 47.33 & 51.83 & 54.86 \\
    EvoPrompt(GA)-OPTS(US) & 49.12 & 53.08 & 56.45 \\
    EvoPrompt(GA)-OPTS(TS) & 50.80 & 53.32 & 56.34 \\
    \bottomrule
    \end{tabular}
  }
  \caption{\label{table:variance-GA}
    Averages of the minimum, median, and maximum scores in three trials with GA-based EvoPrompt. 
  }
\end{table*}

\section{Analysis of Score Variance}
\label{sec:score_var}
We analyze the score variance of EvoPrompt-\ours and EvoPrompt. Tables \ref{table:main-DE} and \ref{table:main-GA} show that EvoPrompt-\ours has a higher variance compared to EvoPrompt. The cause is that OPTS sometimes finds significantly superior prompts in some tasks. Tables \ref{table:variance-DE} and \ref{table:variance-GA} show the average (in 27 tasks) of the minimum, median, and maximum scores in three trials. We can see that the averages of the minimum scores of EvoPrompt-OPTS(TS) match the averages of the maximum scores of EvoPrompt. This implies that EvoPrompt-OPTS(TS) usually finds better or comparative solutions than EvoPrompt.

\section{Discussion of Applicability to Other Tasks}
We discuss the applicability of EvoPrompt-\ours to tasks other than BBH. 
Basically, similar to BBH, the proposed methods can be applied to other tasks like sentiment classification, MedQA, etc., while several modifications will be needed regarding the initial prompts and regular expressions used to extract the answer part.
For instance, in initial prompts for sentiment classification, the user should provide descriptions of the task, such as ``Classify the given sentence as either positive or negative.''
Also, in order to successfully extract answers, providing few-shot examples and modifying regular expressions will be needed.
As long as we can extract the answer part from the response of the task-solving LLM and compare it with the correct answer, we believe that it is not difficult to apply our method to other tasks.

\section{License for Artifacts}
BIG-Bench Hard and lm-evaluation-harness are licensed under the MIT License.
Llama-3-8B-Instruct is licensed under META LLAMA 3 COMMUNITY LICENSE AGREEMENT.
Our code is also released under the MIT license at \url{https://github.com/shiralab/OPTS}.

\section{Artifact Use Consistent with Intended Use}
We declare that we have used the BIG-Bench Hard dataset and Llama-3-8B-Instruct in accordance with their original intended use.
% Additionally, our code we will be releasing should not be used to optimize prompts for generating malicious content, except for research purposes.
Additionally, we prohibit the use of the code we release for optimizing prompts to generate malicious content, except for research purposes.

\section{Experimental Environment}
Our experiments were conducted on a computer running Ubuntu 22.04 with an AMD EPYC 7502P CPU and an NVIDIA A100 GPU, and on another computer running Ubuntu 22.04 with an AMD EPYC 7702P CPU and an NVIDIA A100 GPU.
We used openai 1.40.8 as the python library to access GPT-4o mini, and vllm 0.6.3.post1 as the python library to access Llama-3-8B-Instruct.

\end{document}